%% file: root.tex
\documentclass{article}

\usepackage[final]{corl_2019} %
\usepackage{amssymb}
\usepackage{amsmath}
\usepackage{mathtools} %
\usepackage{graphicx}
\usepackage{subcaption}
\newdimen\figrasterwd
\figrasterwd\textwidth

\usepackage{amsthm}

\usepackage{algorithm}
\usepackage{algpseudocode} %
\algtext*{EndWhile} %
\algtext*{EndIf} %
\algtext*{EndFor} %
\algtext*{EndProcedure}

\newcommand{\est}[1]{\smash{\hat{#1}}}
\newcommand{\safety}{\lambda}
\newcommand{\confidence}{\gamma}
\newcommand{\cautious}{\text{caut}}
\newcommand{\optimistic}{\text{opt}}
\newcommand{\failed}{\textbf{\textit{failed}}}
\newtheorem{definition}{Definition}
\newtheorem{theorem}{Theorem}
\newtheorem{lemma}{Lemma}

\algnewcommand\Title[1]{\item[\textbf{Algorithm:}] \textsuperscript{#1}}
\algnewcommand\Input[1]{\State \textbf{Input:} #1}
\algnewcommand\Initialize[1]{\State \textbf{initialize} #1}
\algnewcommand\Break{\State\textbf{break}}
\algnewcommand\Continue{\State\textbf{continue}}
\algrenewcommand\Return{\State\textbf{return}~}

\DeclareMathOperator*{\argmax}{argmax}

\title{A Learnable Safety Measure}

\author{
  Steve Heim\textsuperscript{1,}\thanks{Equally contributing.}\enspace, Alexander von Rohr\textsuperscript{2,3,}\footnotemark[1]\enspace, Sebastian Trimpe\textsuperscript{2} and Alexander Badri-Spr{\"o}witz\textsuperscript{1} \\
\textsuperscript{1}: Dynamic Locomotion Group, Max Planck Institute for Intelligent Systems, Germany\\
\textsuperscript{2}: Intelligent Control Systems Group, Max Planck Institute for Intelligent Systems, Germany\\
\textsuperscript{3}: Ingenieurgesellschaft Auto und Verkehr (IAV), Germany
}

\usepackage{layouts}
\begin{document}
\maketitle
\begin{abstract}
Failures are challenging for learning to control physical systems since they risk damage, time-consuming resets, and often provide little gradient information.
Adding safety constraints to exploration typically requires a lot of prior knowledge and domain expertise.
We present a safety measure which implicitly captures how the system dynamics relate to a set of failure states.
Not only can this measure be used as a safety function, but also to directly compute the set of safe state-action pairs. Further, we show a model-free approach to learn this measure by active sampling using Gaussian processes.
While safety can only be guaranteed after learning the safety measure, we show that failures can already be greatly reduced by using the estimated measure during learning.
\end{abstract}

\section{Introduction}
Learning control directly on hardware has great promise: learning would enable robots to adapt to changing environments, exploit un-modeled dynamics, as well as greatly decrease the engineering effort required to deploy a robot in the field. 
One of the challenges during the exploration process is that the system might visit failure states, such as a flying robot crashing or a legged robot falling over. 
These failure states can be costly in terms of time, damage to the robot or environment, and are often uninformative for the learning process. 
Unfortunately, the learning agent may not know a priori which actions lead to failure states. Furthermore, there may be actions which lead to unviable states: states which have not yet failed, but from which it is inevitable to reach a failure state in the future. Ideally, the learning agent only explores actions which keep the system within the set of viable states, also known as the \emph{viability kernel}~\cite{aubin2011viability}. \par
Although algorithms to compute conservative approximations of the viability kernel are available, they are contingent on accurate dynamics models, require substantial engineering effort and do not scale well for many types of systems~\cite{bansal2017hamilton, liniger2017real}. 
Alternatively, it can be useful to have a safety function which indicates how close the system is to leaving the viability kernel. 
Safety functions can help guide exploration to stay within the viability kernel without having to know its precise bounds.
However, designing these functions is non-trivial, and faces the same issues commonly seen in designing reward functions: they are typically only approximate indicators of potential failures, require much handcrafting, and often introduce unwanted designer bias into the exploration. \par
We propose a model-free approach to learn a safety function, which captures the notion of viability without requiring the viability kernel to be explicitly computed. 
Our first contribution is a safety measure taken over the set of viable state-action pairs. Intuitively speaking, this measure describes the quantity of actions available that can avoid leaving the viability kernel. It therefore implicitly captures the structure of the system’s dynamics, and how this relates to failure states, making it an effective safety function. 
Our second contribution is an algorithm for model-free learning of probabilistic estimates of both the measure and the viable set, using a Gaussian process (GP).
On the one hand, making no model assumptions means we cannot guarantee safety until the measure has converged.
We show, nonetheless, that the estimated measure can already be used during learning to reduce the number of visits to failure states significantly.
On the other hand, keeping assumptions to a minimum allows this approach to be applied more readily to systems with arbitrary dynamics, where accurate models may be difficult to come by. 
This makes our approach particularly well-suited to systems that are difficult to model and where failures are costly but not critical.
\subsection{Background and Related Work}
\textbf{Safe Learning with Viability Kernels and Back-reachable Sets.} 
Two common model-based approaches to find safe sets are the computation of viability kernels~\citep{aubin2011viability,liniger2017real,wieber2008viability,bayen2002guaranteed} and back-reachable sets~\citep{bansal2017hamilton, piovan2015reachability}. For viability, the user first defines a set of failure states; the viability kernel is then the set of states from which there exist actions, such that the system can avoid the failure set for all time. For back-reachability, a target set is defined; the back-reachable set is the set of states from which there exist actions, such that the system can reach the target set in finite time. In practice, these sets often coincide~\cite{bayen2002guaranteed, zaytsev2018boundaries, kaynama2012computing} and can be used interchangeably\footnote{This is the case when the viability kernel is ergodic.}. 
This depends, however, not only on the system dynamics but also on the specified failure and target sets. For example, the failure set for a walking robot may be defined as all states from which the robot cannot move its center of mass (e.g., it has fallen over and cannot recover), and the target set may be defined as reaching a specific location. If obstacles are blocking the path to the target location, the robot may be outside the back-reachable set of the target set, even though it is inside the viability kernel. \par
There are several algorithms used to compute back-reachable sets or viability kernels in state space; their effectiveness depends greatly on the assumptions used to model the system.
For an overview, we recommend~\citep{bansal2017hamilton, liniger2017real}. 
To circumvent the difficulty of obtaining an accurate model from first-principles, models can also be learned from data.
For example,~\citet{akametalu2014reachability} and~\citet{fisac2018general} learn a GP model of the dynamics of the system and disturbances, and use this to compute a conservative reachable set. 
As the system explores this set, the GP model is refined, and the set can (usually) be expanded. 
\citet{fisac2018general} demonstrate their approach on quadcopter flight. 
They also point out the strong interdependence of safety and learning the system’s true dynamics: safety guarantees are only as good as the models they are based on.
In contrast, we do not model or learn the system dynamics, but a safety measure instead. 
We then estimate the set of viable state-actions directly from our measure, which enables model-free safe learning. 
Although this loses safety guarantees while learning the safety measure, it can be substantially easier to apply to complex systems. \par
Recently, the notion of viability has been extended to sets in state-action space.~\citet{zaytsev2018boundaries} use this to directly link the reachable and viable sets.
\citet{heim2019beyond} use this to quantify the influence of system design on robustness to noisy actions, which is particularly relevant in learning control.
We use the same notion of viability in state-action space, but extend the binary notion of viability (a state-action pair either belongs to the set or not) with a measure.\par
\textbf{Bayesian Optimization and Reinforcement Learning with Safety Functions.}
Recently, safe Bayesian optimization (BO) using GPs has been used to apply model-free learning of controller parameters for systems with failure conditions~\cite{berkenkamp2016safe, schreiter2015safe, schillinger2018safe}. 
In addition to modeling the controller performance, a second GP is used to model the safety of the controller parameter space. 
The safety model is used to restrict active sampling to parameters with a high probability of being safe. 
Though the controller parameters are applied to dynamical systems, safety is evaluated as purely dependent on parameter space, such that it can be considered as a static bandit problem. Thus, each sample of the parameter-space does not affect the safety of future samples. This approach is challenging to apply to situations that include non-steady-state behavior or where a set of controller parameters may be safe for some states but not others.
In contrast to safe BO, we consider the more general case where safety is dependent on the current state. This emphasizes the role of the system dynamics, as they constrain the paths that can be taken through the search space. This type of problem can be modeled as a non-ergodic Markov decision process: that is, where not every state can be reached from every other state. \par
\citet{turchetta2016safe} have extended safe BO to Markov decision processes, and they demonstrate this on a non-ergodic grid-world example, where there exist states which are reachable, but from which the system cannot return. 
The notion of safety as ergodicity was previously formalized by~\citet{moldovan2012safemdp} in the general reinforcement learning context, who also point out the counter-intuitive result that more cautious exploration can often lead to faster convergence. \par
In all of these approaches, it is assumed that a safety function can be sampled whenever visiting a new point in the search space (whether this is the parameter or state space). 
Safety is then inferred for nearby, unvisited points.
The probable safety of these states can then be guaranteed using certain assumptions on the safety function, such as Lipschitz-continuity. 
However, this safety function is typically user-defined, and only indicative of what might cause failure. 
For example,~\citet{schillinger2018safe} use the temperature of the engine at steady-state,~\citet{berkenkamp2016safe} use a minimum performance threshold, and~\citet{turchetta2016safe} use the ground inclination a rover needs to negotiate. 
Just as guarantees for model-based methods depend on the quality of the model, safe BO depends on a well-chosen safety function. In practice, the safety function is often chosen to be more conservative than strictly necessary. 
In contrast, our safety measure implicitly encodes the structure of the system dynamics and a definition of failure states. 
Furthermore, we show that the measure does not need to be known a priori, and can be learned in a model-free manner by sampling.
With no model assumptions, safety guarantees can only be given once the measure has converged. 
Prior knowledge can, however, be introduced to reduce failures significantly.\par
The rest of the paper is structured as follows: In Section \ref{sec:problem}, we define all the necessary objects and introduce the safety measure. These concepts are illustrated with a toy example. In Section \ref{sec:probabilistic}, we extend this to a probabilistic setting and present an algorithm to learn the safety measure in a model-free context. In Section \ref{sec:results}, we show simulation results using our algorithm and point out key properties. In Section \ref{sec:conclusion} we summarize our contribution and potential future work.
\section{A Measure over the Viable Set}\label{sec:problem}
\begin{figure}[bth]
    \centering
    \includegraphics[width=0.8\textwidth]{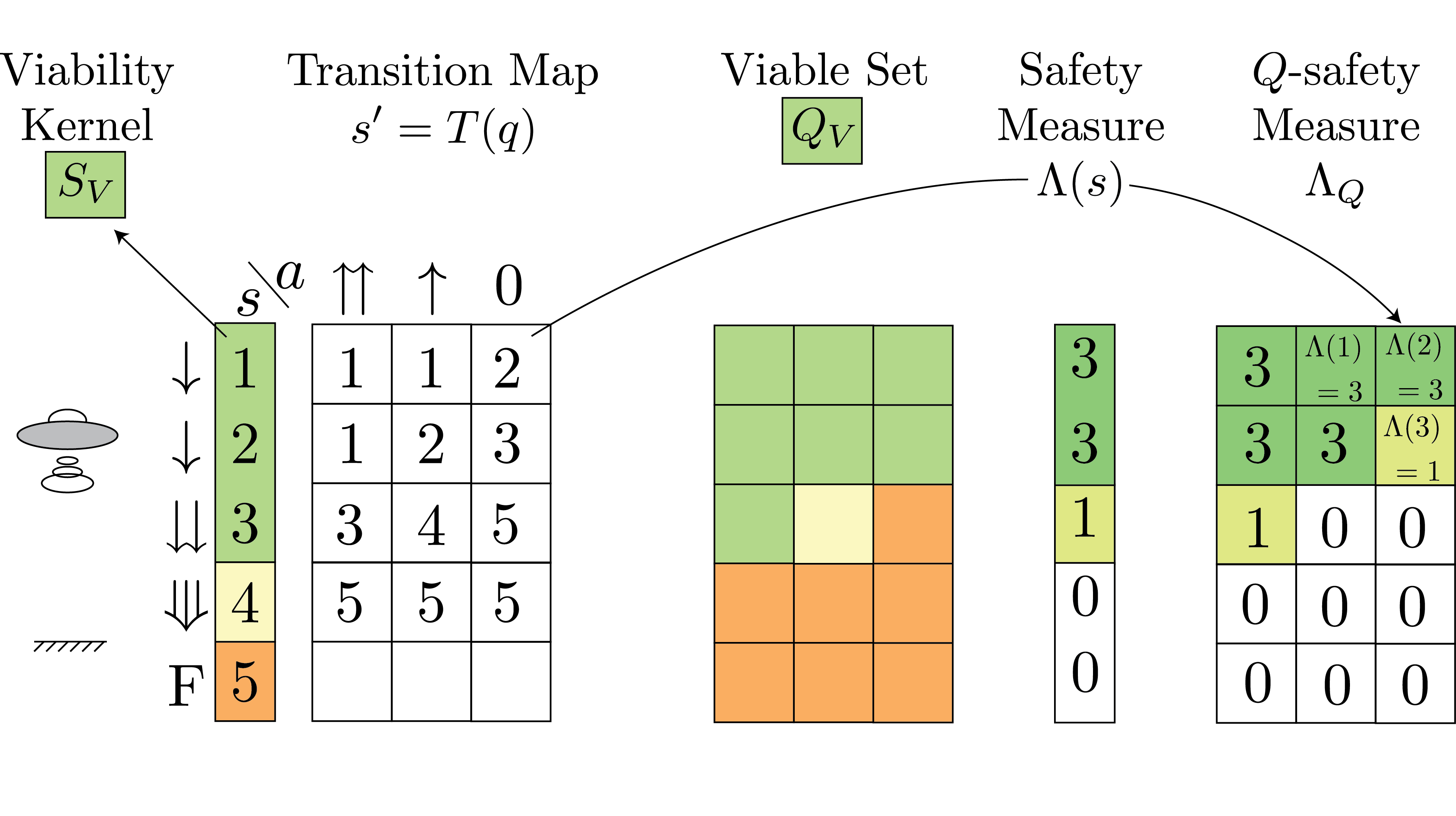}
    \caption{Shown are the transition map of our toy model, as well as each object: the viability kernel $S_V$, the viable set $Q_V$, the safety measure $\Lambda$ and the $Q$-safety measure $\Lambda_Q$. Both $S_V$ and $Q_V$ are highlighted in green. We also highlight state-action pairs which result directly in failure in orange, and those that are unviable in yellow. The arrows and illustration are only to help with intuition.}
    \label{fig:toy_model}
\end{figure}

We consider systems with deterministic dynamics of the form $ s' =T(s, a)$, where $ s \in S$ is a state, $a \in A$ is an action, and $T$ is the transition map of the dynamics to a new state $s' $. 
The set of failure states $S_F \subset S$ can be defined arbitrarily. For the sake of simplicity, we consider here a set of states from which there are no meaningful transitions and the system would need to be reset or replaced.
We will define objects in the state-action space $Q \coloneqq S \times A$.
We use the shorthand $s'=T(q)$ where $q \coloneqq \left(s, a\right) \in Q$. We will illustrate the defined objects on a discrete grid-world, amenable to pen-and-paper computation, and shown in Fig. \ref{fig:toy_model}. \par
\textbf{Toy Model.} Intuitively, the transition map in Fig.\ \ref{fig:toy_model} can be thought of as representing a hovering spaceship affected by gravity, which is stronger near the ground. 
The spaceship can apply two levels of thrusters or allow itself to fall.
The failure set is $S_F: \{5\}$, when the spaceship crashes. \par
We next define important mathematical objects for this work and illustrate them with the toy example. First, we define the viability kernel $S_V$.
\begin{definition}[Viability Kernel]
The \emph{viability kernel} $S_V \subset S \setminus S_F$ is the maximal set of all states $s \in S$, from which there exists an action that keeps the system inside $S_V$  (cf.~\citep[Chapter~1.1]{aubin2011viability}). 
\end{definition}
By its definition, states outside of $S_V$ have either already failed, or will fail within finite time~\citep{aubin2011viability}.
In the toy model, the viability kernel is $S_V = \{ 1, 2, 3\}$: for each of these states, there exists at least one action which can keep the spaceship from ever failing. Avoiding failure does not require ergodicity: the state $s=3$ is viable, but it can no longer reach the other viable states.
Note also that $s=4$ is neither in the viability kernel $S_V$ nor in the set of failure states $S_F$: it has not yet failed but cannot avoid reaching the failure set eventually. Next, we define the viable set in state-action space, $Q_V$.
\begin{definition}[Viable Set]
The \emph{viable set} $Q_V \subset Q$ is the maximal set of all state-actions $q$, such that $s'=T(q) \in S_V$.
\end{definition}
By its definition, the viability kernel $S_V$ is the projection of the viable set $Q_V$ onto state space: for any state in $S_V$, the agent can sample a state-action from $Q_V$ which maps back into itself~\citep{heim2019beyond}. Both $S_V$ and $Q_V$ are highlighted with green in Fig.~\ref{fig:toy_model}. We can now define the safety measure $\Lambda$.
\begin{definition}[Safety Measure]
The \emph{safety measure} $\Lambda$ is the $n$-dimensional volume of the viable set $Q_V$.
When applied to a point $ s \in S$, $\Lambda(s) \in \mathbb{R}_{\geq 0}$ is the measure of the corresponding slice of $Q_V$.
\end{definition}
We use the Lebesgue measure for continuous spaces (assuming the sets are Lebesgue-measurable), and the counting measure for discrete spaces.
Intuitively, a higher value $\Lambda(s)$ indicates that at state $s$, more viability-maintaining actions are available. 
A low value indicates that the agent should be very precise and deliberate since very few actions allow the system to avoid failure.
In our toy model, for example, $\Lambda(3) = 1$ means that there is only one action which allows the system to avoid failure, this step and in the future.
We can now map $\Lambda$ into state-action space with the transition matrix.
\begin{definition}[$Q$-Safety Measure]
The \emph{$Q$-Safety Measure} $\Lambda_Q$ is defined as $\Lambda(s')$ where $s' = T(q)$. We use the shorthand $\Lambda_Q(q) = \Lambda(s')$.
\end{definition}
Next, we define safe level sets as the sets with measure $\Lambda_Q > \safety$, where the minimum safety level $\safety$ is a non-negative scalar.
\begin{definition}[Safe Level Sets]
A safe level set $S_\safety$ is a set of states where $\Lambda(s) > \safety$. A safe level set $Q_\safety$ is a set of state-action pairs which map into $S_\safety$, such that $\Lambda_Q(q) > \safety$.
\end{definition}

In other words, sampling from $Q_\safety$ will map the system to a state from which there is at least one action which maintains a safety level $\safety$. Thus, the system can continue to choose actions which maintain a minimum safety level of $\safety$ indefinitely. In Fig.~\ref{fig:toy_model}, the safe level-sets $Q_{\safety=0}$ and $Q_{\safety=2}$ are highlighted in different shades of green. This can be useful when certain types of disturbances are expected. 
We can recover the viable set from $\Lambda_Q$ with $Q_V = Q_{\safety=0}$, since viability implies $\Lambda(s) > 0$. 
Thus, if the safety measure $\Lambda_Q$ is known, both the viable set $Q_V$ and $\Lambda$ can be obtained directly. 
If only $\Lambda$ is known, $\Lambda_Q$ can be computed directly using the transition map $T$.
In the next section, we will use these facts to learn $\Lambda_Q$ in a model-free fashion by sampling the dynamics.

\section{Learning the Measure by Sampling}\label{sec:probabilistic}

Given a system with an unknown transition map $T$ and an unknown failure set $S_F$, our main objective is to estimate the viable set $\est{Q}_V$ as a large, conservative approximation of the true viable set, $\est{Q}_V \subseteq Q_V$.
Since we assume an accurate dynamics model is unavailable, we directly sample the transition map $T$ from a given state $s$ by choosing an action $a$.
Specifically, we begin sampling sequences from an initial state $s$.
We then receive the tuple $\left(s',\, \failed\right)$, where $s' = T(q)$ is the new state, and $\failed$ is a boolean indicating if $s' \in S_F$. The estimate $\est{\Lambda}_Q$ can only be updated with the estimate $\est{\Lambda}$, except when sampling a failure.
The goal is to choose actions $a$ that are informative for learning the safety measure while avoiding the failure set $S_F$.
\par
To achieve this goal, we model $\Lambda_Q$, from which we compute $\est{Q}_V$ and $\est{\Lambda}$.
The estimate $\est{\Lambda}_Q$ can already be used during learning to avoid actions with a low estimated probability of being safe.

\par

\subsection{Convergence Properties}\label{sub:convergence}
To examine the requirements for $\est{\Lambda}_Q$ to converge to the true measure $\Lambda_Q$, we separately consider the viable set $Q_V$ and its complement $Q_V^c$, the set of unviable and failed state-action pairs.
\begin{theorem}\label{th:convergence}
 Under the assumption of infinite random sampling over $Q$, the measure $\est{\Lambda}_Q$ converges to the correct value $0$ for all $q \in Q_V^c$.
\end{theorem}
\addtocounter{theorem}{-1}
A proof for discrete state-action spaces can be found in appendix \ref{app:proof}.
Once the measure $\est{\Lambda}_Q$ inside $Q_V^c$ has converged, the estimate for the viable set $\est{Q}_V$ is tightly bounded from above. Therefore, the estimated safety measure is also tightly bounded to be  $\est{\Lambda} \leq \Lambda$.
We can then ensure $\est{\Lambda}_Q$ converges to the true measure by initializing optimistically, such that initially $\est{\Lambda} \geq \Lambda$. 
These two conditions of infinite sampling and optimistic starts are typical for model-free learning~\citep[Chapter~2.6]{sutton2018book} but are also impractical. 
In particular, optimistic starts encourage visits to the failure set. 
We will now extend $\Lambda_Q$ to a probabilistic setting, and use confidence bounds over the measure to estimate $\est{Q}_V$.
Although this loses the guarantee of converging to the true $\Lambda_Q$, we show that in practice it allows us to converge to conservative subsets while reducing failures effectively.

\subsection{Probabilistic Estimates: Modeling $\Lambda_Q$ with Gaussian processes}

A probabilistic estimate allows us to (i) include prior knowledge without an explicit model of the dynamics, and
(ii) estimate the uncertainty of the safety measure for a given state-action pair $q$, which we will use for active sampling.
When modeling $q$ as a random variable, the distribution should only allow for positive values and also have non-zero probability mass on the point zero, to model the probability of a point being unviable.
We use a normal distribution as a practical approximation, where the probability mass below zero is treated as the discrete probability for the point zero.
Specifically, we use a GP~\citep{rasmussen2006gaussian} to model the probabilistic estimate of $\Lambda_Q$. 
The posterior estimate of the measure at any point in $Q$ is normally distributed, and it includes the prior assumptions on the estimate as well as the samples $\mathcal{D} = (q_i,\est{\Lambda}_i(s_i'))$,
\begin{align*}
    \est{\Lambda}_{Q}(q)|\mathcal{D} \sim \mathcal{N}(\mu(q), \sigma^2(q))
\end{align*}
where $\est{\Lambda}_{Q}(q)|\mathcal{D}$ means the estimate is conditioned on the samples, $\mathcal{N}$ is the normal distribution, $\mu$ is the posterior mean function and $\sigma^2$ the posterior variance, given by the covariance function.
The prior mean and covariance function can be used to encode the prior knowledge of the measure function, such as smoothness or known safe sets.\par

Given $\est{\Lambda}_Q$, the probability that a state-action pair belongs to the safe level set $Q_{\safety}$ can be calculated using the cumulative distribution function of the normal distribution $F_{\est{\Lambda}_Q}$ as
\begin{equation*}
     \mathbb{P}\left[\est{\Lambda}_Q|\mathcal{D} > \safety\right] \approx 1 - F_{\est{\Lambda}_Q}\left[\safety\right].
\end{equation*}

\subsection{A Learning Algorithm for $\Lambda_Q$}

We provide an approach for learning $\est{\Lambda}_Q$ and the derived $\est{Q}_V$ and $\est{\Lambda}$, described in Algorithm~\ref{alg:learning}.
As discussed in Section \ref{sub:convergence}, convergence requires an \emph{optimistic} estimate of $Q_V$, such that the intitial estimate $\est{\Lambda} \geq \Lambda$. Otherwise, a viable state-action pair may be incorrectly assigned the value $0$.
At the same time, the algorithm should use a \emph{cautious} estimate for active sampling to reduce the probability of failing. 
To deal with this challenge, we use an optimistic set $\est{Q}_\optimistic$ to compute $\est{\Lambda}$. A separate, cautious set $\est{Q}_\cautious$ is used for active sampling. We obtain these as
\begin{align*}
    \est{Q}_\optimistic(\confidence_\optimistic) & = \begin{cases}
    1 & \text{if } \mathbb{P}\left[\est{\Lambda}_Q|\mathcal{D} > 0\right] > \confidence_\optimistic \\
    0 & \text{otherwise},\\
    \end{cases} \\
    \est{Q}_\cautious(\confidence_\cautious, \safety_\cautious) & = \begin{cases}
    1 & \text{if }\mathbb{P}\left[\est{\Lambda}_Q|\mathcal{D} > \safety_\cautious \right] > \confidence_\cautious \\
    0 & \text{otherwise}\\
    \end{cases}
\end{align*}
by thresholding the probability with a minimum confidence $\confidence \in \left[0, 1\right]$.
The algorithm has three tuning parameters: $\confidence_\optimistic$ governs the level of optimism in $\est{Q}_\optimistic$, and $\safety_\cautious$ and $\confidence_\cautious$ govern the level of caution for active sampling. 
Choosing $\confidence_\cautious \geq \confidence_\optimistic$ ensures that $\est{Q}_\cautious \subseteq \est{Q}_\optimistic$, so we never purposefully explore outside the current estimate of the viability set. The algorithm samples the action from the cautious set $\est{Q}_\cautious$ with highest variance. By actively reducing variance, the confidence in the measure is increased. Choosing actions with high variance also encourages exploration of the state space.
\setlength{\textfloatsep}{4pt}
\begin{algorithm}[t]
    \caption{Learning the safety measure} \label{alg:learning}
    \begin{algorithmic}[1]
        \Input{initial measure estimate~$\est{\Lambda}_Q$; thresholds $\confidence_\cautious$, $\confidence_\optimistic$, and $\safety_\cautious$; initial state $s_0$; maximum number of samples $n$.}
        \While{$i < n$}
            \State Compute $\est{\Lambda}$, $\est{Q}_\optimistic$ and $\est{Q}_\cautious$ from $\est{\Lambda}_Q$.
            \State $A_{caut} \gets \forall a$ s.t. $(s_i,a) \in \est{Q}_\cautious$. \Comment{\parbox[t]{0.3\linewidth}{Determine safe actions.}}
            \If {$A_{caut}$ is empty} 
                \State $a_i \gets \argmax\limits _{(s_i,a) \in A} \mathbb{P}[a \in \est{Q}_\cautious]$ \Comment{\parbox[t]{0.3\linewidth}{Take safest action.}}
            \Else
                \State $a_i  \gets \argmax\limits _{a \in A_{caut}} \sigma^2(s_i,a)$. \Comment{\parbox[t]{0.3\linewidth}{Explore based on variance of the GP model.}}
            \EndIf
            \State $(s_{i+1},\, \failed) \gets T(s_i,a_i)$. \Comment{\parbox[t]{0.3\linewidth}{Sample the dynamics.}}
            \If {\failed} 
                \State Update $\mathcal{D}$ with $((s_i,a_i),\, 0)$ and recompute $\est{\Lambda}_Q$ 
                \State $s_{i+1} \gets$ random state from $\est{Q}_\cautious$. \Comment{\parbox[t]{0.3\linewidth}{Reset if failed.}}
            \Else
                \State Compute $\est{\Lambda}(s_{i+1})$ from $\est{Q}_\optimistic$
                \State Update $\mathcal{D}$ with $((s_i,a_i), \est{\Lambda}(s_{i+1}))$ and recompute $\est{\Lambda}_Q$ 
                \Comment{\parbox[t]{0.3\linewidth}{Update measure estimate.}}
            \EndIf
        \EndWhile
    \end{algorithmic}
\end{algorithm}

\section{Results}\label{sec:results}

We have tested our algorithm in simulation, and provide a Python implementation using the SciPy~\cite{scipy2001} and GPy packages~\cite{gpy2014}. 
The code can is available in the supplementary material and online at \url{github.com/sheim/vibly}, and includes example code to reproduce the results shown here, some additional examples, and a template to implement dynamics of other systems.
\par
We report the results of two examples, which each highlight a specific challenge: 
dealing with unviable state-action pairs, 
and dealing with complex dynamics.
We also use the second example to suggest guidelines for choosing the algorithm parameters, though this will typically be system-specific.
Both examples are low-dimensional, and the ground-truth is computed by brute force.
This allows us to easily choose reasonable parameters for the GP model, which is otherwise a separate challenge for using Gaussian processes.
In practice, choosing these parameters is highly system-dependent \cite{rai2017bayesian}.
We use a covariance function from the Mat\'{e}rn family~\citep[Chapter~4]{rasmussen2006gaussian}, which has two parameters: the length scales for each input dimension and the signal variance.
The length scales describe how fast the measure changes when moving away from a known state-action pair.
The second parameter is the signal variance, which relates to the total variation of the measure estimate $\est{\Lambda}_Q$. Details for the models are in the appendix. \par
\begin{figure*}[t!]
    \centering
    \begin{subfigure}[b]{0.49\textwidth}
        \centering
        \includegraphics[width=1\textwidth]{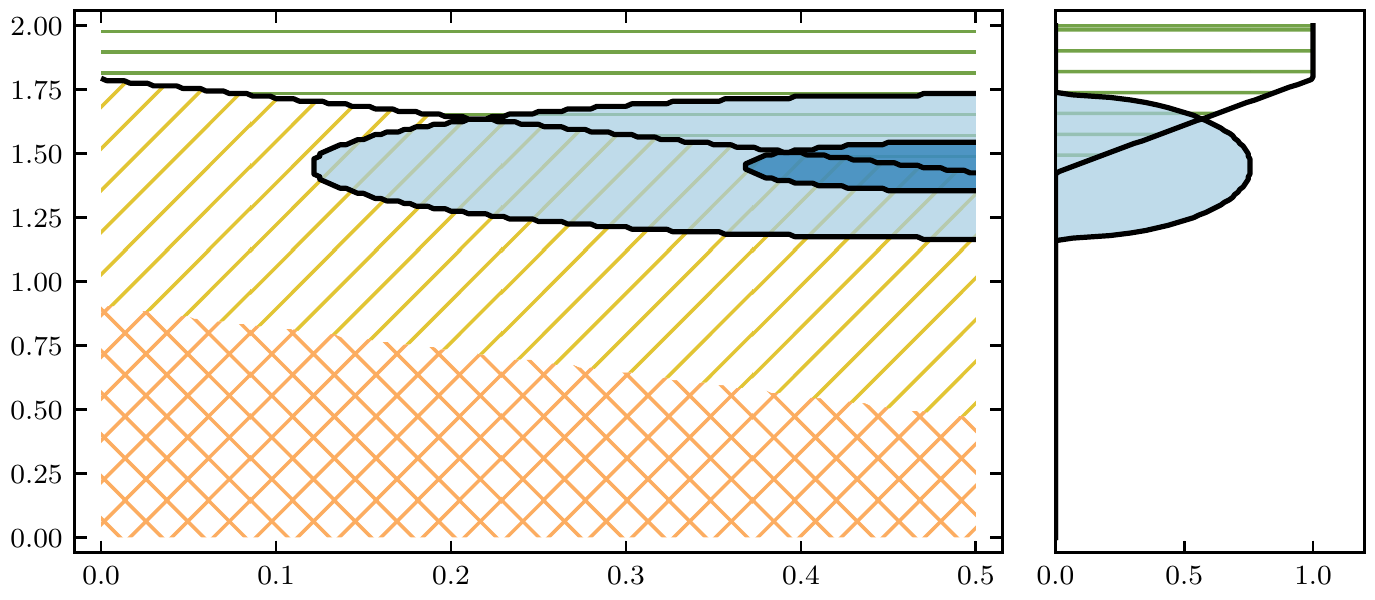}
        \caption{Prior}\label{subfig:hover_a}
    \end{subfigure}%
    ~ 
    \begin{subfigure}[b]{0.49\textwidth}
        \centering
        \includegraphics[width=1\textwidth]{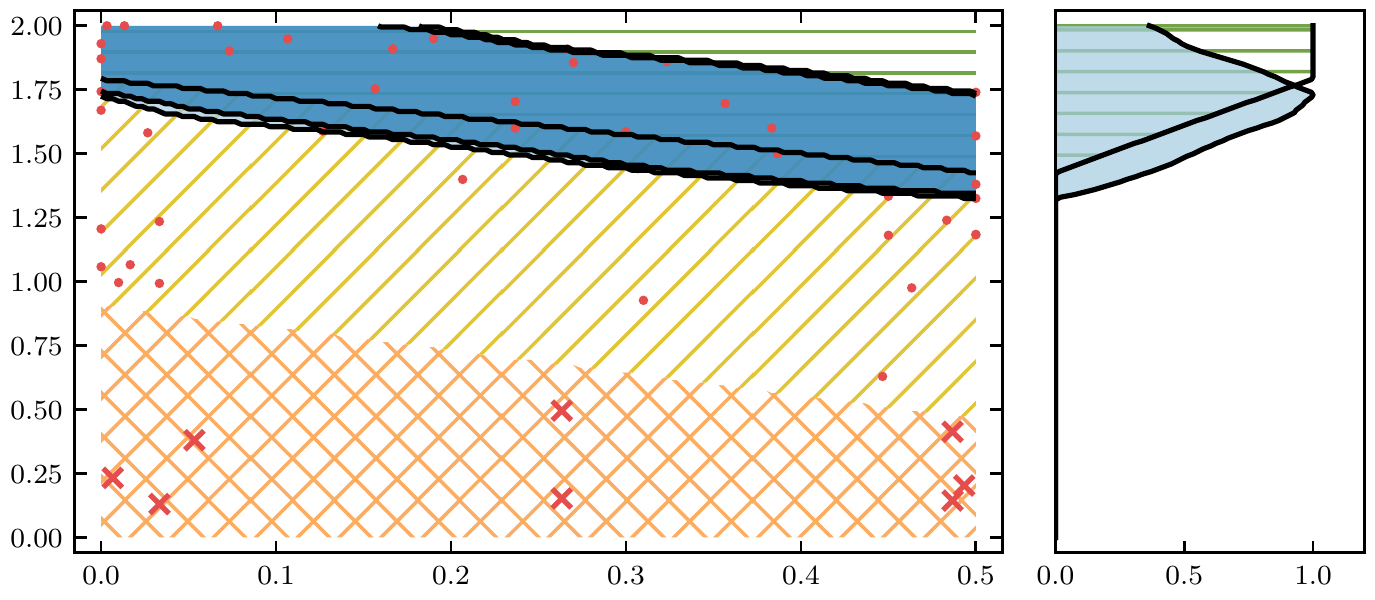}
        \caption{50 samples}\label{subfig:hover_b}
    \end{subfigure}\\
    \begin{subfigure}[b]{1\textwidth}
        \centering
        \includegraphics[width=1\textwidth]{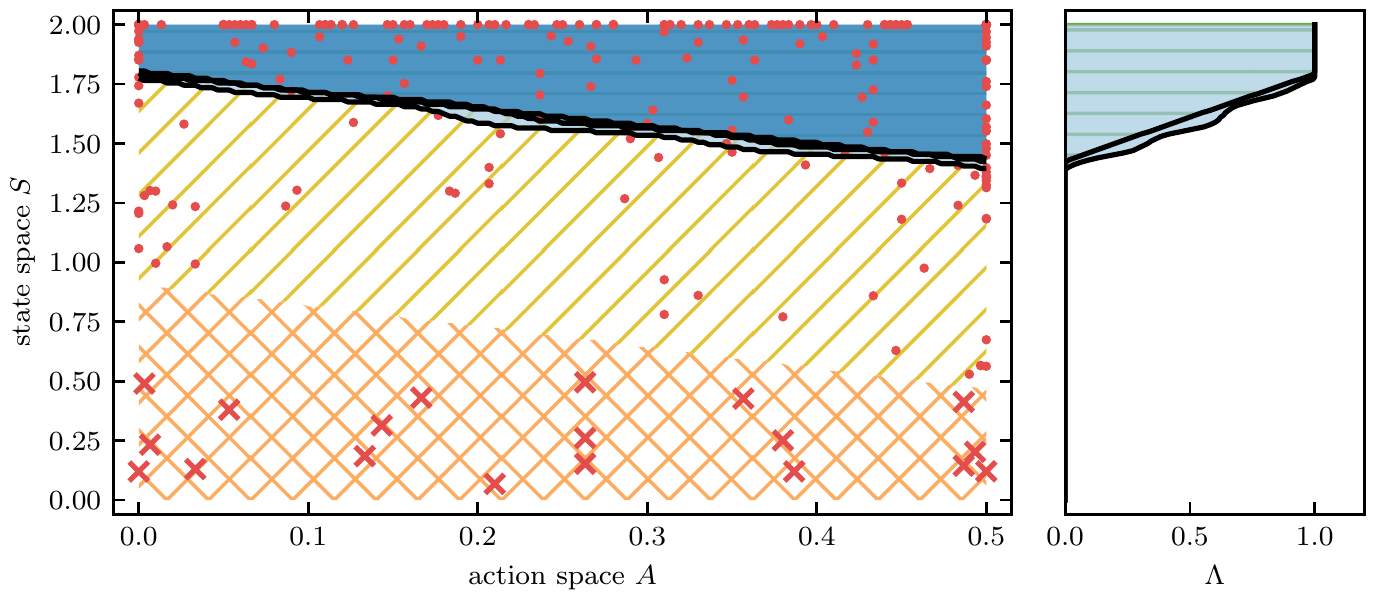}
        \caption{250 samples}\label{subfig:hover_c}
    \end{subfigure}\\
    \begin{subfigure}[b]{1\textwidth}
    \includegraphics[width=1\textwidth]{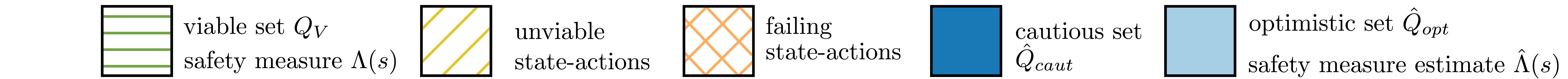}
    \end{subfigure}
    \caption{The progress of learning the $\est{\Lambda}_Q$ for the spaceship model. Starting from an inaccurate prior \ref{subfig:hover_a}, after 50 samples \ref{subfig:hover_b} and after 250 samples \ref{subfig:hover_c}.
    Red dots indicate samples, red crosses samples that result in failures.
    After 250 samples, the optimistic set (light blue) and the measure $\Lambda$ (green) are close to the ground truth.
    The cautious set $\est{Q}_\cautious$ (dark blue) begins with a substantial region outside the viable set (\ref{subfig:hover_c}), but quickly converges to the ground truth.
    }\label{fig:hover}
\end{figure*}

\textbf{Unviable state-action pairs.}
Our first example is based on the hovership spaceship from Section \ref{sec:problem}.
The model has been modified with a continuous state-action space, and the dynamics have been adjusted to increase the portion of the state-action space which is unviable. The GP prior mean is purposefully initialized poorly, such that most of the initial estimate $\est{Q}_\optimistic$ lies outside the true $Q_V$.
This example shows that the algorithm can cope with unviable states, even though the ground truth is only sampled at failure. The confidence thresholds $\confidence_\optimistic$, $\confidence_\cautious$ and $\safety$ are initialized to encourage rapid initial exploration, then gradually increased to speed up convergence to a safe subset of $Q_V$.
After 250 samples, it has nearly converged to the ground-truth, with an 8\% failure rate (see Fig. \ref{subfig:hover_c}). 
In both examples shown in this paper, the confidence thresholds are increased linearly with each iteration as a heuristic that helps speed up convergence and reduces failures.\par

\begin{figure*}[t!]
    \centering
    \begin{subfigure}[b]{0.49\textwidth}
        \centering
        \includegraphics[width=1\textwidth]{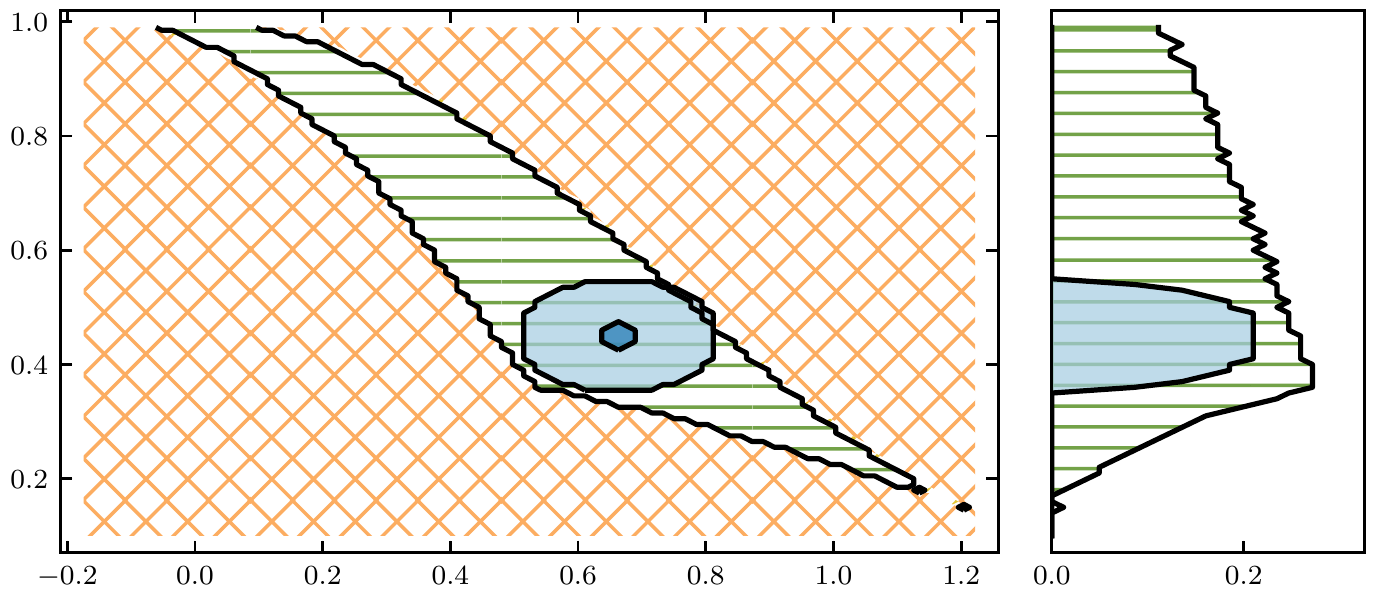}
        \caption{prior} \label{subfig:slip1}
    \end{subfigure} 
    ~ 
    \begin{subfigure}[b]{0.49\textwidth}
        \centering
        \includegraphics[width=1\textwidth]{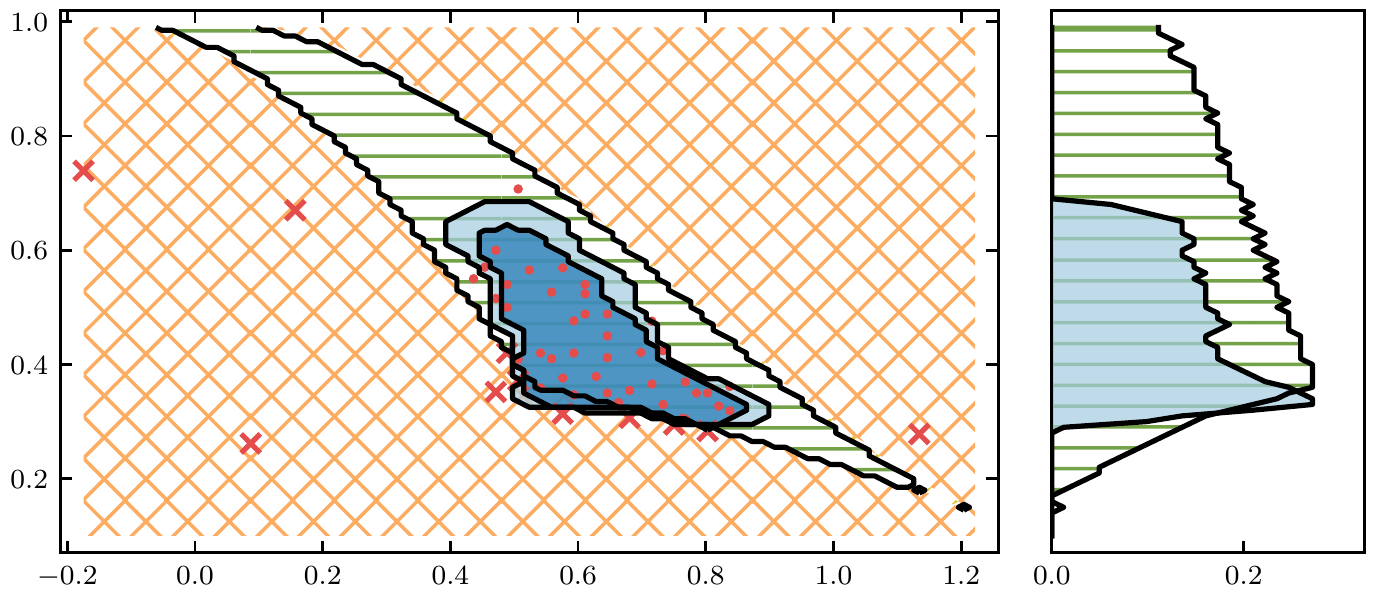}
        \caption{50 samples}\label{subfig:slip2}
    \end{subfigure}\\
    \begin{subfigure}[b]{1\textwidth}
        \centering
        \includegraphics[width=1\textwidth]{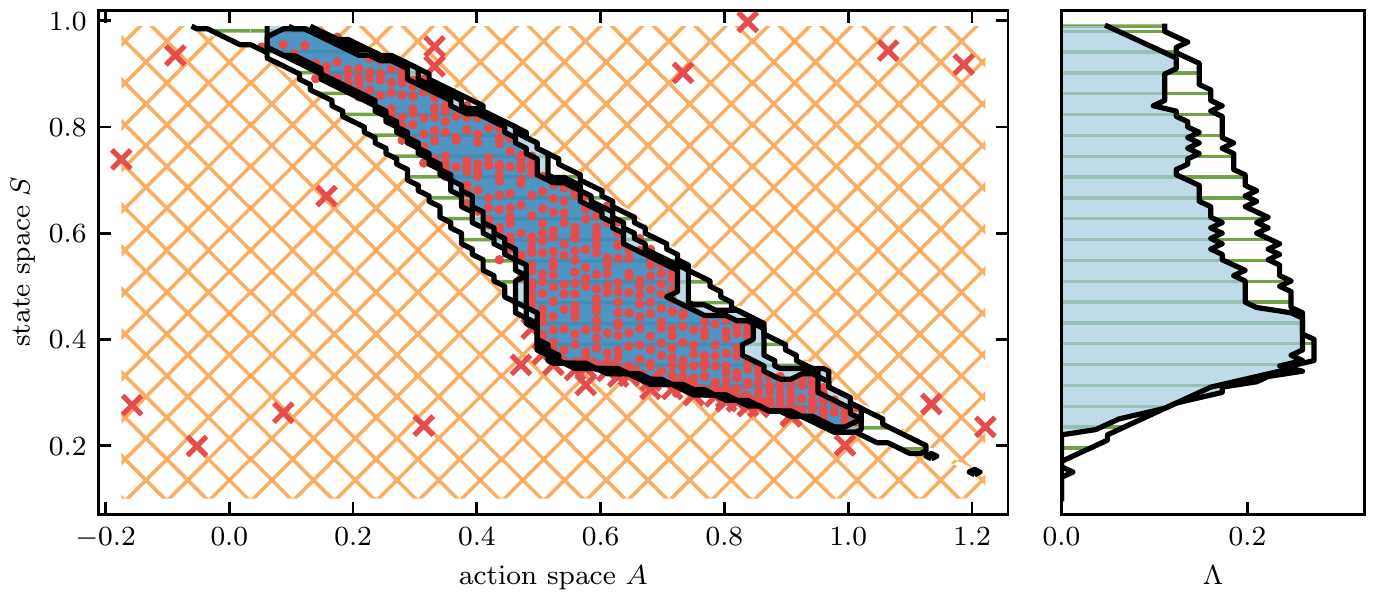}
        \caption{500 samples}\label{subfig:slip3}
    \end{subfigure}\\
    \begin{subfigure}[b]{1\textwidth}
    \includegraphics[width=1\textwidth]{graphics/leggenda.png}
    \end{subfigure}
    \caption{Learning the viable set and the corresponding measure $\est{\Lambda}$ for a SLIP model. Starting from a conservative prior (\ref{subfig:slip1}), after 50 samples (\ref{subfig:slip2}) and after 500 samples (\ref{subfig:slip3}).
    The color coding is as in Fig.~\ref{fig:hover}. A non-smooth infeasibility constraint bounds the bottom edge of the viable set. This violates the smoothness assumption of the GP, and requires many samples to learn accurately. The other edges are learned fairly accurately without many failures sampled. Actions close to the left edge are avoided, as they bring the system to states with low safety measure.
    }\label{fig:slip}
\end{figure*}

\textbf{Complex, unmodeled dynamics.}
Our second example is a simulation of the spring-loaded inverted pendulum (SLIP) model, a low-dimensional idealized model commonly used to design controllers for running robots~\citep{piovan2015reachability, wensing2013high, hubicki2016atrias}.
Control, and therefore learning, is applied once per step-cycle, at the apex of the flight phase.
The system dynamics are therefore treated as a nonlinear, discrete map with a 2-dimensional state-action space; this nonlinear map is obtained by numerically simulating the full dynamics between two apex events.
The set of failures, which includes falling and reversing direction, is evaluated on the full state space of the continuous dynamics.
For this system, the measure $\Lambda_Q$ has a non-smooth edge on the lower part of the state space, due to an infeasibility\footnote{Infeasible state-action pairs have no physical meaning and cannot exist, such as starting underground.} constraint (see Fig. \ref{fig:slip}).
Attempting to sample infeasible state-action pairs returns a failure.
At this discontinuity, the smoothness assumptions encoded in the GP are violated. Therefore, more failures are sampled to learn the border of the viable set (see Fig. \ref{subfig:slip1}).
At the other borders of the viable set, where the smoothness assumption holds, the estimated sets approach the border despite sampling very few to no failures.
When getting closer to the border of the viable set, the measure shrinks and the border of the set can be inferred without sampling unviable state-action pairs. \par
We also use this example to illustrate best practices for a realistic scenario, and the influence of different choices for the tuning parameters $\confidence_\optimistic$, $\confidence_\cautious$ and $\safety$.
The prior covariance function is obtained from simulations of an incorrect model, in which spring constant of the SLIP model is 20\% lower. 
Since the covariance function encodes qualitative properties, it is reasonable to use a low fidelity simulation to obtain these GP parameters.
The GP prior mean is typically more sensitive to simulation inaccuracies.
It is therefore chosen around a known operating point, which we assume can be determined with conventional means without the need for a full model.
Ideal operating points will feature a stable equilibrium-point or slow divergence from the operating point.
Thus, the learning system can drive down variance locally before exploring more distant states, and the confidence bound $\confidence_\optimistic$ and $\confidence_\cautious$ can be initialized more aggressively.
We initialize the operating point of the SLIP model near a known limit-cycle of the running model.
Although the prior is very conservative (see Fig. \ref{fig:slip}), the algorithm converges to a conservative yet nearly maximal approximation of the viable set $Q_V$, with a failure rate of 8\% after 500 samples.

\section{Discussion and Outlook}\label{sec:conclusion}
The first contribution of this paper is a safety measure taken over the set of viable state-action pairs. 
While this measure is useful in itself, computing viable sets relies on accurate models and is often intractable for systems with complex, high-dimensional dynamics. Our second contribution is a probabilistic, model-free approach to learn this measure and a safe set of state-action pairs using GPs.
On the one hand, this makes it applicable to a variety of systems. 
On the other hand, making almost no assumptions means there are no hard guarantees for avoiding failure, even with a reasonable prior.
This approach is therefore appropriate for systems which are difficult to model and where failures are costly but not critical, such as robots with soft or compliant components~\citep{surovik2018any, Buchler18ControlMusculo} and small to mid-sized legged robots~\citep{rai2017bayesian, xie2018feedback, hwangbo2019learning}. \par

An issue with our current algorithm is that old samples contain old, potentially incorrect estimates of the measure, which can interfere with newer samples.
A principled approach to only keep informative samples would improve the estimate and reduce computation costs.
As with most other learning approaches, scaling to higher dimensions is a key challenge.
An exploration strategy with an information-theoretic approach, especially with a heteroscedastic model (with state-dependent uncertainty), should improve both accuracy as well as sample-efficiency.
In practice, it may be desirable to balance information gain of the safety measure and performance. How to balance this in a principled manner is an open question.
We believe that leveraging a dynamics model will be key to scaling. How to map assumptions of the dynamics to the safety model requires further investigation.
In addition, sample-efficiency might be greatly improved by updating all state-action pairs that are close in a dynamical sense instead of a Euclidean sense.
How to obtain such a metric of closeness in state-action space is a problem we find is both challenging and has significant potential.

\clearpage
\acknowledgments{We thank Dominik Baumann, Matthias Neumann-Brosig and Friedrich Solowjow for insightful discussions during the project and while preparing the manuscript, as well as the anonymous reviewers for constructive and thorough feedback. We thank IMPRS-IS for the academic development of Steve Heim and Alexander von Rohr. This work was partly funded through the Cyber Valley Initiative and the Max Planck Society.}

\bibliography{library}  %
\clearpage
\input{appendix.tex}

\end{document}

%% file: appendix.tex
\appendix
\section{Convergence of the measure estimate for unviable state-action pairs}\label{app:proof}
We will show here the convergence result for systems with discrete time and finite, discrete state-action spaces.
Specifically we show that, under the assumption of inifinite sampling, the estimated measure $\hat{\Lambda}_Q$ for the unviable state-action pairs $q \in Q_V^c$ converges to the true value of $0$ when following the updates

\begin{equation} \label{eq:update}
\est{\Lambda}_Q(q) = \begin{cases} 0 & \text{if } \textbf{\textit{failed}}\\
\est{\Lambda}(s')& \text{else.}
\end{cases}
\end{equation}
A direct consequence is that the estimated measure $\est{\Lambda}$ for all unviable states also converges to the true value of 0. This provides an upper bound for the measure. We proceed to the theorem.

\begin{theorem}
 Under the assumption of infinite random sampling over $Q$, the measure $\est{\Lambda}_Q$ converges to the correct value $0$ for all $q \in Q_V^c$.
\end{theorem}

We define $S_U = \left( S_V \cup S_F \right)^c$ as the set of unviable states, and $Q_U \coloneqq S_U \times A$. We also define the operator $\operatorname{len}(s)$, which returns the integer length of the longest trajectory starting from the state $s$.
We begin by showing that this theorem holds for all $q \in Q_U$, which ensures that the estimated measure converges to $0$ for all $s \in S_U$.
Consider all possible trajectories starting from any state $s \in S_U$. By the definition of viability, they are all \emph{acyclic}, meaning no state is ever visited more than once. They therefore all end in $S_F$ within finite time.
\begin{lemma}
The longest trajectory starting from any state $s \in S_U$ has length $\operatorname{len}(s) \leq n$, where $n$ is the number of states in $S_U$.
\end{lemma}
This can be proven by contradiction. Let us assume that the longest trajectory starting in $S_U$ has length $n_{longest} > n$. We take a sub-trajectory of length $n$; due to the acyclicity condition, this trajectory has has visited $n$ unique states, and therefore has visited all states in $S_U$. It therefore cannot be lengthened without breaking the acyclicity condition, contradicting our assumption.
\begin{lemma}
For every $i = 1, \ldots, n_{longest}$, there exists at least one state $s \in S_U$, for which the longest trajectory beginning from that state has length $i$.
\end{lemma}
Again, this can be shown by contradiction. Let us assume there are no states with $\text{len}(s) = 1$. Then take the longest trajectory starting from any $s \in S_U$, and proceed to the last state in the trajectory. By our assumption, $\operatorname{len}(s)>1$, which implies that there is at least one action available from this state which would avoid failure and therefore increase the length of the trajectory by at least 1, contradicting the our previous statement. This reasoning can be extended to all other $i$ up to $n_{longest}$ by inserting shorter states in the failure set $S_F$ and repeating this process. \par
\begin{proof}
Now it is clear that sampling any $q$ from a state with $\operatorname{len}(s) = 1$ will immediately transition to $S_F$, and therefore will be updated with the ground truth as per \ref{eq:update}. Once each of such $q$ has been sampled once, the measure estimate will be $0$ for any state with $\operatorname{len}(s) = 1$. At this point, sampling any $q$ from a state with $\operatorname{len}(s) = 2$ will be updated with 0, and so on until all $s \in S_U$ have converged to 0. \par
We can now turn our attention to the remaining $q \in Q_V^c$. By definition, these are state-action pairs that transition to $S_U$ in a single step. Therefore, as soon as the estimated measure for all $s \in S_U$ has converged to 0, these will also be updated correctly with $\est{\Lambda}(s) = 0$.
\end{proof}
\clearpage
\section{Dynamics of Simulated Examples}\label{app:dynamics}
We include here the details of the dynamics for the systems used in Section \ref{sec:results}. The implementation in Python is available in the supplementary material. We also include in the supplementary material code to compute the true viable sets, although this is only computationally tractable for small systems.

\subsection{Hovering Spaceship} \label{sub:hover}

This example is a hovering spaceship loosely based on the toy example in Section \ref{sec:problem}, with a continuous state-action space. The spaceship has a single state, a vertical position, and is affected by nonlinear gravity. The non-constant gravity has been adjusted to accentuate the issue of nonviable which have not yet failed. The dynamics are:
\begin{equation}
    \dot{s} = g_0 + \tanh{\left(0.75\, s\right)}g - a
\end{equation}
where $s$ is the height, $g_0$ is a baseline gravitational acceleration, and $g$ is a coefficient for the gravity which increases with state. The spaceship can counteract gravity with the action $0 \leq a \leq a_{max}$. The failure set is defined as $s \geq s_{max}$. We also model a control frequency $\omega$, such that the spaceship can only choose a new thrust $a$ once every $\frac{1}{\omega}$ seconds. This control delay further accentuates the unviable states. The reader is encouraged to adjust these parameters in the code to see how this affects both the viable sets, and the learning of the safety measure. \\
The parameters used to generate the graph in the paper are:

\begin{center}
\begin{tabular}{|l l l l|} 
\hline
base gravitational acceleration &$g_0$ :& 0.1 & \\
\hline
gravitational acceleration &$g$ :& 1 & \\
\hline
max thrust &$a_{max}$ :& 0.5 & \\
\hline
ground height &$s_{max}$:& 2 & \\
\hline
control frequency &$\omega$ :& 1 & \\ 
\hline
\end{tabular}
\end{center}

\subsection{Spring Loaded Inverted Pendulum}

The spring-loaded inverted pendulum is a common model for understanding running dynamics, both in biomechanics and robotics. The body is represented by a point-mass, and a massless spring represents the leg. It has hybrid dynamics, meaning the governing equations of motion switch between different phases, and cyclic orbits. We begin each cycle at the apex, the highest point during flight phase, when vertical velocity is zero. Each step cycle has 3 phases: A flight phase terminating with a touchdown event, a stance phase terminating with a liftoff event, and a second flight phase terminating with an apex event. \\
Between two apex events, we integrate the full state $\left[x, y, \dot{x}, \dot{y}\right]^\intercal$, where $x$ and $y$ are the horizontal and vertical positions of the point-mass. During each flight phase, the dynamics are

\begin{equation*}
\left[
\begin{array}{c}
\ddot{x}\\
\ddot{y}\\
\end{array}
\right] = \left[
\begin{array}{c}
0\\
-g\\
\end{array}
\right],
\end{equation*}
where $g$ is the gravitational acceleration. During stance phase, the dynamics are
\begin{align*}
\left[
\begin{array}{c}
\ddot{x}\\
\ddot{y}\\
\end{array}
\right] & = \frac{k\left(l_0-l\right)}{m}\left[
\begin{array}{c}
\sin\left(\theta\right)\\
\cos\left(\theta\right)\\
\end{array}
\right] - 
\left[
\begin{array}{c}
0 \\
g\\
\end{array}
\right] \\
\theta & = \arctan2\left(\frac{y}{x}\right) - \frac{\pi}{2} \\
l & = \sqrt{(x^2 + y^2)},
\end{align*}
where $k$ is the spring stiffness, $l_0$ is the spring resting length, and $m$ is the mass. For concise notation, the reference frame is centered on the foot during stance. In the implementation, the foot position is also tracked and accounted for. The events are detected with
\begin{align*}
\text{touchdown: } & l = l_0 \\
\text{liftoff: } & \theta = \arctan2\left(\frac{y}{x}\right) - \frac{\pi}{2} \\
\text{apex: } & \dot{y} = 0.
\end{align*}

For simplicity, we can examine the state once per step cycle, on the so-called Poincar{\'e} section, at the apex of flight. At apex, all the potential energy at apex is contained in the height of the point-mass, and the kinetic energy in the forward velocity (the vertical velocity is zero by the definition of apex).
Since the system is energy-conservative, we can use the potential energy normalized by total energy as our single state.
We thus use as state  $s = \frac{E_{\text{pot}}}{E_{\text{pot}} + E_{\text{kin}}} = \frac{g y}{\frac{\dot{x}^2}{2} + g y}$, where $E_{\text{pot}}$ and $E_\text{kin}$ are the potential and kinetic energy, respectively.
For our simulations, we also define a single action: the landing angle of attack of the leg, $a = \alpha$, which is instantly reset to the desired angle at each apex. \\

For the figures in the paper, we used the following parameters, which are similar to human averages:
\begin{center}
\begin{tabular}{|l l l l|} 
\hline
gravitational acceleration &$g$ :& 9.81 &$\left[m/s^2\right]$ \\
\hline
body mass &$m$ :& 80 &$\left[kg\right]$ \\
\hline
spring stiffness &$k$ :& 8200 &$\left[N/m\right]$ \\ 
\hline
spring resting length &$l_0$:& 1 &$[m]$ \\
\hline
\end{tabular}
\end{center}

%% file: root.bbl
\begin{thebibliography}{27}
\providecommand{\natexlab}[1]{#1}
\providecommand{\url}[1]{\texttt{#1}}
\expandafter\ifx\csname urlstyle\endcsname\relax
  \providecommand{\doi}[1]{doi: #1}\else
  \providecommand{\doi}{doi: \begingroup \urlstyle{rm}\Url}\fi

\bibitem[Aubin et~al.(2011)Aubin, Bayen, and Saint-Pierre]{aubin2011viability}
J.-P. Aubin, A.~M. Bayen, and P.~Saint-Pierre.
\newblock \emph{Viability theory: new directions}.
\newblock Springer Science \& Business Media, 2011.

\bibitem[{Bansal} et~al.(2017){Bansal}, {Chen}, {Herbert}, and
  {Tomlin}]{bansal2017hamilton}
S.~{Bansal}, M.~{Chen}, S.~{Herbert}, and C.~J. {Tomlin}.
\newblock Hamilton-jacobi reachability: A brief overview and recent advances.
\newblock In \emph{2017 IEEE 56th Annual Conference on Decision and Control
  (CDC)}, pages 2242--2253, Dec 2017.
\newblock \doi{10.1109/CDC.2017.8263977}.

\bibitem[{Liniger} and {Lygeros}(2019)]{liniger2017real}
A.~{Liniger} and J.~{Lygeros}.
\newblock Real-time control for autonomous racing based on viability theory.
\newblock \emph{IEEE Transactions on Control Systems Technology}, 27\penalty0
  (2):\penalty0 464--478, March 2019.
\newblock \doi{10.1109/TCST.2017.2772903}.

\bibitem[{Wieber}(2008)]{wieber2008viability}
P.~{Wieber}.
\newblock Viability and predictive control for safe locomotion.
\newblock In \emph{2008 IEEE/RSJ International Conference on Intelligent Robots
  and Systems}, pages 1103--1108, Sep. 2008.
\newblock \doi{10.1109/IROS.2008.4651022}.

\bibitem[Bayen et~al.(2002)Bayen, Cr{\"u}ck, and Tomlin]{bayen2002guaranteed}
A.~M. Bayen, E.~Cr{\"u}ck, and C.~J. Tomlin.
\newblock Guaranteed overapproximations of unsafe sets for continuous and
  hybrid systems: Solving the hamilton-jacobi equation using viability
  techniques.
\newblock In \emph{Hybrid Systems: Computation and Control}, pages 90--104.
  Springer Berlin Heidelberg, 2002.

\bibitem[Piovan and Byl(2015)]{piovan2015reachability}
G.~Piovan and K.~Byl.
\newblock Reachability-based control for the active slip model.
\newblock \emph{The International Journal of Robotics Research}, 34\penalty0
  (3):\penalty0 270--287, 2015.
\newblock \doi{10.1177/0278364914552112}.

\bibitem[{Zaytsev} et~al.(2018){Zaytsev}, {Wolfslag}, and
  {Ruina}]{zaytsev2018boundaries}
P.~{Zaytsev}, W.~{Wolfslag}, and A.~{Ruina}.
\newblock The boundaries of walking stability: Viability and controllability of
  simple models.
\newblock \emph{IEEE Transactions on Robotics}, 34\penalty0 (2):\penalty0
  336--352, April 2018.
\newblock \doi{10.1109/TRO.2017.2782818}.

\bibitem[Kaynama et~al.(2012)Kaynama, Maidens, Oishi, Mitchell, and
  Dumont]{kaynama2012computing}
S.~Kaynama, J.~Maidens, M.~Oishi, I.~M. Mitchell, and G.~A. Dumont.
\newblock Computing the viability kernel using maximal reachable sets.
\newblock In \emph{Proceedings of the 15th ACM International Conference on
  Hybrid Systems: Computation and Control}, HSCC '12, pages 55--64, New York,
  NY, USA, 2012. ACM.
\newblock \doi{10.1145/2185632.2185644}.

\bibitem[{Akametalu} et~al.(2014){Akametalu}, {Fisac}, {Gillula}, {Kaynama},
  {Zeilinger}, and {Tomlin}]{akametalu2014reachability}
A.~K. {Akametalu}, J.~F. {Fisac}, J.~H. {Gillula}, S.~{Kaynama}, M.~N.
  {Zeilinger}, and C.~J. {Tomlin}.
\newblock Reachability-based safe learning with gaussian processes.
\newblock In \emph{53rd IEEE Conference on Decision and Control}, pages
  1424--1431, Dec 2014.
\newblock \doi{10.1109/CDC.2014.7039601}.

\bibitem[{Fisac} et~al.(2019){Fisac}, {Akametalu}, {Zeilinger}, {Kaynama},
  {Gillula}, and {Tomlin}]{fisac2018general}
J.~F. {Fisac}, A.~K. {Akametalu}, M.~N. {Zeilinger}, S.~{Kaynama},
  J.~{Gillula}, and C.~J. {Tomlin}.
\newblock A general safety framework for learning-based control in uncertain
  robotic systems.
\newblock \emph{IEEE Transactions on Automatic Control}, 64\penalty0
  (7):\penalty0 2737--2752, July 2019.
\newblock \doi{10.1109/TAC.2018.2876389}.

\bibitem[{Heim} and {Spr\"{o}witz}(2019)]{heim2019beyond}
S.~{Heim} and A.~{Spr\"{o}witz}.
\newblock Beyond basins of attraction: Quantifying robustness of natural
  dynamics.
\newblock \emph{IEEE Transactions on Robotics}, 35\penalty0 (4):\penalty0
  939--952, Aug 2019.
\newblock \doi{10.1109/TRO.2019.2910739}.

\bibitem[{Berkenkamp} et~al.(2016){Berkenkamp}, {Schoellig}, and
  {Krause}]{berkenkamp2016safe}
F.~{Berkenkamp}, A.~P. {Schoellig}, and A.~{Krause}.
\newblock Safe controller optimization for quadrotors with gaussian processes.
\newblock In \emph{2016 IEEE International Conference on Robotics and
  Automation (ICRA)}, pages 491--496, May 2016.
\newblock \doi{10.1109/ICRA.2016.7487170}.

\bibitem[Schreiter et~al.(2015)Schreiter, Nguyen-Tuong, Eberts, Bischoff,
  Markert, and Toussaint]{schreiter2015safe}
J.~Schreiter, D.~Nguyen-Tuong, M.~Eberts, B.~Bischoff, H.~Markert, and
  M.~Toussaint.
\newblock Safe exploration for active learning with gaussian processes.
\newblock In \emph{Machine Learning and Knowledge Discovery in Databases},
  pages 133--149. Springer International Publishing, 2015.

\bibitem[Schillinger et~al.(2016)Schillinger, Ortelt, Hartmann, Schreiter,
  Meister, Nguyen-Tuong, and Nelles]{schillinger2018safe}
M.~Schillinger, B.~Ortelt, B.~Hartmann, J.~Schreiter, M.~Meister,
  D.~Nguyen-Tuong, and O.~Nelles.
\newblock Safe active learning of a high pressure fuel supply system.
\newblock In \emph{Proceedings of The 9th EUROSIM Congress on Modelling and
  Simulation}, pages 286--292, 2016.

\bibitem[Turchetta et~al.(2016)Turchetta, Berkenkamp, and
  Krause]{turchetta2016safe}
M.~Turchetta, F.~Berkenkamp, and A.~Krause.
\newblock Safe exploration in finite markov decision processes with gaussian
  processes.
\newblock In \emph{Advances in Neural Information Processing Systems 29}, pages
  4312--4320. 2016.

\bibitem[Moldovan and Abbeel(2012)]{moldovan2012safemdp}
T.~M. Moldovan and P.~Abbeel.
\newblock Safe exploration in markov decision processes.
\newblock In J.~Langford and J.~Pineau, editors, \emph{Proceedings of the 29th
  International Conference on Machine Learning (ICML-12)}, ICML '12, pages
  1711--1718, July 2012.

\bibitem[Sutton and Barto(2019)]{sutton2018book}
R.~S. Sutton and A.~G. Barto.
\newblock \emph{Reinforcement learning: An introduction, 2nd Edition},
  volume~1.
\newblock {MIT} {press} {Cambridge}, 2019.
\newblock URL \url{http://incompleteideas.net/book/the-book-2nd.html}.

\bibitem[Rasmussen and Williams(2006)]{rasmussen2006gaussian}
C.~Rasmussen and C.~Williams.
\newblock \emph{Gaussian Processes for Machine Learning}.
\newblock MIT Press, 2006.

\bibitem[Jones et~al.(2001--)Jones, Oliphant, Peterson, et~al.]{scipy2001}
E.~Jones, T.~Oliphant, P.~Peterson, et~al.
\newblock {SciPy}: Open source scientific tools for {Python}, 2001--.
\newblock URL \url{http://www.scipy.org/}.
\newblock [Online; accessed 23.09.2019].

\bibitem[{GPy}(since 2012)]{gpy2014}
{GPy}.
\newblock {GPy}: A gaussian process framework in python.
\newblock \url{http://github.com/SheffieldML/GPy}, since 2012.

\bibitem[{Rai} et~al.(2018){Rai}, {Antonova}, {Song}, {Martin}, {Geyer}, and
  {Atkeson}]{rai2017bayesian}
A.~{Rai}, R.~{Antonova}, S.~{Song}, W.~{Martin}, H.~{Geyer}, and C.~{Atkeson}.
\newblock Bayesian optimization using domain knowledge on the atrias biped.
\newblock In \emph{2018 IEEE International Conference on Robotics and
  Automation (ICRA)}, pages 1771--1778, May 2018.
\newblock \doi{10.1109/ICRA.2018.8461237}.

\bibitem[{Wensing} and {Orin}(2013)]{wensing2013high}
P.~M. {Wensing} and D.~E. {Orin}.
\newblock High-speed humanoid running through control with a 3d-slip model.
\newblock In \emph{2013 IEEE/RSJ International Conference on Intelligent Robots
  and Systems}, pages 5134--5140, Nov 2013.
\newblock \doi{10.1109/IROS.2013.6697099}.

\bibitem[Hubicki et~al.(2016)Hubicki, Grimes, Jones, Renjewski, Spr{\"o}witz,
  Abate, and Hurst]{hubicki2016atrias}
C.~Hubicki, J.~Grimes, M.~Jones, D.~Renjewski, A.~Spr{\"o}witz, A.~Abate, and
  J.~Hurst.
\newblock Atrias: Design and validation of a tether-free 3d-capable spring-mass
  bipedal robot.
\newblock \emph{The International Journal of Robotics Research (IJRR)},
  35\penalty0 (12):\penalty0 1497--1521, 2016.
\newblock \doi{10.1177/0278364916648388}.

\bibitem[Surovik et~al.(2019)Surovik, Wang, Vespignani, Bruce, and
  Bekris]{surovik2018any}
D.~Surovik, K.~Wang, M.~Vespignani, J.~Bruce, and K.~E. Bekris.
\newblock Adaptive tensegrity locomotion: Controlling a compliant icosahedron
  with symmetry-reduced reinforcement learning.
\newblock \emph{International Journal of Robotics Research (IJRR)}, 2019.

\bibitem[{B{\"u}chler} et~al.(2018){B{\"u}chler}, {Calandra}, {Sch{\"o}lkopf},
  and {Peters}]{Buchler18ControlMusculo}
D.~{B{\"u}chler}, R.~{Calandra}, B.~{Sch{\"o}lkopf}, and J.~{Peters}.
\newblock Control of musculoskeletal systems using learned dynamics models.
\newblock \emph{IEEE Robotics and Automation Letters}, 3\penalty0 (4):\penalty0
  3161--3168, Oct 2018.
\newblock \doi{10.1109/LRA.2018.2849601}.

\bibitem[{Xie} et~al.(2018){Xie}, {Berseth}, {Clary}, {Hurst}, and {van de
  Panne}]{xie2018feedback}
Z.~{Xie}, G.~{Berseth}, P.~{Clary}, J.~{Hurst}, and M.~{van de Panne}.
\newblock Feedback control for cassie with deep reinforcement learning.
\newblock In \emph{2018 IEEE/RSJ International Conference on Intelligent Robots
  and Systems (IROS)}, pages 1241--1246, Oct 2018.
\newblock \doi{10.1109/IROS.2018.8593722}.

\bibitem[Hwangbo et~al.(2019)Hwangbo, Lee, Dosovitskiy, Bellicoso, Tsounis,
  Koltun, and Hutter]{hwangbo2019learning}
J.~Hwangbo, J.~Lee, A.~Dosovitskiy, D.~Bellicoso, V.~Tsounis, V.~Koltun, and
  M.~Hutter.
\newblock Learning agile and dynamic motor skills for legged robots.
\newblock \emph{Science Robotics}, 4\penalty0 (26), 2019.
\newblock \doi{10.1126/scirobotics.aau5872}.

\end{thebibliography}
